\documentclass[conference]{IEEEtran}
\usepackage{cite}

\usepackage{amsmath,amssymb,amsfonts}
\usepackage{algorithmic}
\usepackage{graphicx}
\usepackage{textcomp}
\usepackage{xcolor}
\def\BibTeX{{\rm B\kern-.05em{\sc i\kern-.025em b}\kern-.08em
    T\kern-.1667em\lower.7ex\hbox{E}\kern-.125emX}}

\usepackage{caption}	
\usepackage{subcaption}	
\usepackage{amsmath}	
\usepackage{graphicx}	
\usepackage{multirow}

\begin{document}

\title{Diversity and Novelty MasterPrints:\\ Generating Multiple DeepMasterPrints for Increased User Coverage
}

\author{\IEEEauthorblockN{M Charity}
\IEEEauthorblockA{\textit{Tandon Engineering} \\
\textit{New York University}\\
Brooklyn, NY, US \\
mlc761@nyu.edu}
\and
\IEEEauthorblockN{Nasir Memon}
\IEEEauthorblockA{\textit{Tandon Engineering} \\
\textit{New York University}\\
Brooklyn, NY, US \\
memon@nyu.edu}
\and
\IEEEauthorblockN{Zehua Jiang}
\IEEEauthorblockA{\textit{Tandon Engineering} \\
\textit{New York University}\\
Brooklyn, NY, US \\
zj2086@nyu.edu}
\and
\IEEEauthorblockN{Abhi Sen}
\IEEEauthorblockA{\textit{Tandon Engineering} \\
\textit{New York University}\\
Brooklyn, NY, US \\
as15117@nyu.edu}
\and
\IEEEauthorblockN{Julian Togelius}
\IEEEauthorblockA{\textit{Tandon Engineering} \\
\textit{New York University}\\
Brooklyn, NY, US \\
julian@togelius.com}
}

\maketitle

\begin{abstract}
This work expands on previous advancements in genetic fingerprint spoofing via the DeepMasterPrints and introduces Diversity and Novelty MasterPrints. This system uses quality diversity evolutionary algorithms to generate dictionaries of artificial prints with a focus on increasing  coverage of users from the dataset. The Diversity MasterPrints focus on generating solution prints that match with users not covered by previously found prints, and the Novelty MasterPrints explicitly search for prints with more that are farther in user  space than previous prints. Our multi-print search methodologies outperform the singular DeepMasterPrints in both coverage and generalization while maintaining quality of the fingerprint image output.
\end{abstract}


\section{Introduction}

Recent work has demonstrated that user authentication systems based on biometrics can be potentially vulnerable to dictionary attacks~\cite{masterprint}. 
In contrast to well-known spoofing techniques, dictionary attacks do not rely on biometric samples of a targeted individual,  but instead, exploit weaknesses of the specific biometric modality (or its deployment).  In dictionary attacks, biometric samples that tend to match multiple identities are artificially generated. One method for doing this is to  train a model that is able to generate fake biometrics that are representative of the training distribution. Sampling new examples from such trained models can yield  biometric tokens (such as faces or fingerprints) which are sufficiently lifelike to be recognized as real by both humans and biometric authentication systems. Given that no individual token is likely to authorize 100\% of the users, an effective attack needs to use multiple samples that make up the dictionary. 

Most practical biometric-based authentication systems allow multiple trials during verification. This is done to improve usability by decreasing the chance of a false non-match. 
This calls for a need to develop a dictionary of artificially made fingerprints that jointly increase the probability of triggering a false match. The state of the art for generating these fake fingerprints - called MasterPrints \cite{RoyMasterPrint_ICB2018} and its extension DeepMasterPrints \cite{bontrager2018deepmasterprints} - but currently, such these approaches are sub-optimal as the systems only generate a single print to cover a portion of the total set of users. What is needed is a dictionary of DeepMasterPrints which collectively maximizes the probability of a match with k attempts for any random user in a database. 
This work proposes two related heuristics to the problem of generating a DeepMasterPrint dictionary using the generator + evolution setup pioneered by Bontrager et al. \cite{bontrager2018deepmasterprints}. The first solution performs multiple sequential search processes, and for every new search, it modifies the objective metric so as to exclude users covered by previous searches. The second solution uses a novelty search algorithm to find MasterPrints that are sufficiently far from each other. Experimental results show that the constructed dictionaries provide significantly improved performance as opposed to evolving singular Deep Master Prints. 

\section{Background}

\noindent{\bf{Biometric Security:}} It is well known that in spite of its numerous advantages, a fingerprint-based biometric system is potentially  vulnerable to a variety of  attacks \cite{uludag2004attacks}.
Among these, attacks on the input level have received the most attention, since it exploits the standard interaction of the modality and does not require access to the internals of the system.  
Spoofing attacks focus on presenting biometric samples from a targeted individual; these are common and widely studied. From the attacker's perspective, it becomes necessary to collect a representative sample of the victim's biometric features, and generate specific examples to attack the system. Spoofing techniques and defenses vary widely and are actively developed.
\noindent{\bf{MasterPrints and DeepMasterPrints:}} Many consumer electronic devices, such as smartphones, incorporate fingerprint sensors for user authentication. The sensors embedded in these devices are generally small and the resulting images are, therefore, limited in size. To compensate for the limited size, these devices acquire multiple partial impressions that cover  a single finger during enrollment to ensure that at least one of them will successfully match with the image obtained from the user during authentication.
Much work has been done in the biometric domains to create fake samples that can impersonate one or many users. For fingerprint impersonations, Roy generated what is known as “MasterPrints” - fake partial fingerprints created using evolutionary methods with the intention to impersonate one or more users by exploiting the minutiae data of a user’s prints. \cite{roy2017masterprint} \cite{RoyMasterPrint_ICB2018} For this work, we focus on the partial fingerprint domain and expanding on the “DeepMasterPrints” work \cite{bontrager2018deepmasterprints} which evolved latent vectors to produce artificial partial fingerprints that maximize the impersonation rate of as many users as possible.  
This vector is evolved with the objective to generate an image that a verification system (such as the proprietary fingerprint authentication system: VeriFinger) would consider to simultaneously be multiple enrolled users within the training set given a specific FMR. 
However, this method only produced a singular master print and there was no way to account for the missing percentage of users not covered by the fake print. 

\noindent{\bf{Generators:}} Regardless of the domain usage or degree to which they are applied, these artificial biometrics are typically generated via machine learning. A 1-dimensional latent vector is fed into a neural network trained on the impersonation domain dataset (fingerprints, voices, faces, etc.) to output a generated sample that imitates the features displayed by the training data. The latent vector allows for modification on a specific feature (i.e. nose shapes and skin complexion for faces, pitch for voices, and lines arcs for finger prints) by mutating a value in the vector itself in order to produce a change in the sample. 

Most of the previous work with biometric impersonation \cite{shmelkin2021generating,tariq2021real,korshunov2018deepfakes,nguyen2020generating} use some variation of a Generative Adversarial Network (GAN) \cite{goodfellow2014generative} architecture to generate these samples. GANs are trained on the dataset via an adversarial method that determines if the generated sample is fake (generated) or real. While effective, it requires much more fine-tuning and the input latent vector cannot be controlled at a micro-level to alter the features necessary to impersonate a specific user. However, variational autoencoders (VAEs) \cite{kingma2019introduction} are trained to encode the distribution of the features to a vector while still being capable of generating samples via the latent vector input in the same way a GAN does. The training process for VAEs is evaluated with a reconstruction and Kullback-Leibler divergence loss instead of with a discriminator.

\begin{figure}[hbt]
    \centering
    \includegraphics[width=0.8\linewidth]{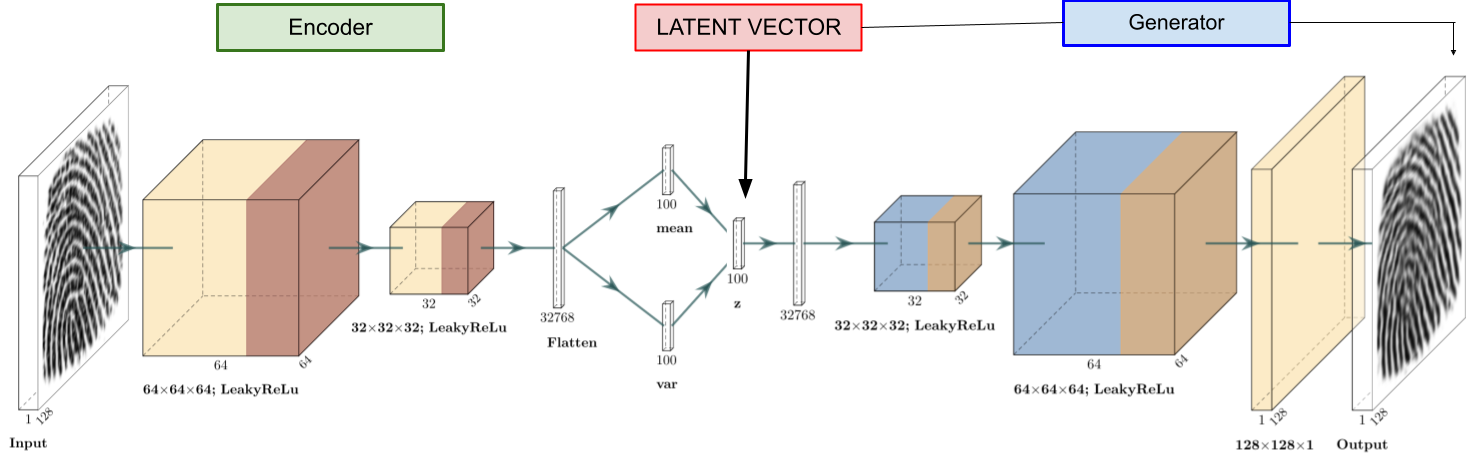}
    \caption{Diagram of the VAE architecture used in the experiment. The generator is the decoder model.}
    \label{fig:vae_arch}
\end{figure}

\noindent{\bf{Evolutionary Search:}} Evolutionary computation and evolutionary search are biological inspired algorithms that foster search through a space via improvement of a population of samples towards an optimization goal called a fitness function \cite{eiben2003introduction}. Typically starting with a randomly initialized set, these samples are mutated and sometimes crossed with other samples from the population until the ideal sample is found or until the overall fitness of the population reaches a certain threshold. 

There exist evolutionary algorithms (given certain assumptions on the search space) that can find a good-enough solution faster \cite{hansen2001completely}, find sets of solutions that vary along certain measures \cite{mouret2015illuminating}, or simply search for more novel solutions \cite{lehman2011novelty}. Bontrager's original work \cite{bontrager2018deepmasterprints} as well as our own uses the Covariance Matrix Adaptation Evolutionary Search (CMA-ES) algorithm - an algorithm that searches based on evolutionary search for difficult non-linear non-convex black-box optimization problems, as described by Hansen \cite{hansen2001completely}. This algorithm assumes that solutions are encoded as vectors of real numbers, and continually updates a probability function from which it samples new solutions. In the setup utilized by Bontrager et al, CMA-ES mutates and evolves samples represented as latent vectors that are passed to the generator model. Building on this concept, our work explores method to find more diverse yet highly-covering samples that can expand on the match coverage of users within the dataset.

\section{Proposed Methods}

\noindent{\bf{Diversity Heuristic:}} Instead of having one generated master fingerprint, as done by Bontrager's DeepMasterPrints technique, we create a dictionary of prints that match as many users as possible. Each generation of prints is created by removing previously matched users from the training pool. This is intended to maximize diversity and coverage for the final set of artificial fingerprints. We call these resulting fingerprints Diversity MasterPrints. An equation for calculating the fitness value for the diversity heuristic is denoted as $diversity\_value = \frac{u_i}{U}$ where $u_i$ is the number of unseen users matched from the sample image $i$ and $U$ is the total number of unseen users left from the entire dataset. 


Like DeepMasterPrints, Diversity MasterPrints are generated via a latent vector and passed through a generator model. The evolutionary fitness function for the algorithm is based on percentage of users covered from the remaining subset pool, which becomes increasingly smaller and easier to find as users are removed. The Diversity MasterPrints are verified against real users via a classification system, just as the DeepMasterPrints were. 
With this method, the generated dictionary may include prints that cover the same users. Even in the generation process, some Diversity MasterPrints can be generated that have overlap with previously found users no longer in the subset user pool. As such, more prints than are necessary may be created via this method. However, the combined user coverage of this dictionary of generated prints is still substantially more than that of the singular DeepMasterPrint - as we will show later in results section of this paper.

\noindent{\bf{Novelty Heuristic:}} Like the diversity heuristic, the novelty heuristic generates MasterPrints to maximize diversity and coverage for as many users but uses novelty search - a divergent search algorithm that focuses on finding samples that are farther away in characteristics from previously made samples while maintaining a minimum fitness \cite{lehman2011novelty}. 
The users are encoded as binary vectors and novelty score is determined by the minimum distance between the generated print's vector and each vector from the current dictionary of prints. We call these resulting fingerprints Novelty MasterPrints. The Novelty MasterPrints are generated the same way as the Diversity MasterPrints; passing an evolved latent vector through a generator and evolving towards the ideal fitness value: a threshold novelty score - which becomes harder to reach after every new sample is added to the dictionary - calculated from the generated print. An equation for the novelty score calculated for each sample is shown in \ref{eqn:novelty_score} where $d$ is the set of matched user vectors from the current saved dictionary stored and $x$ is the current matched user vector from the generated print.
\begin{equation*}
\label{eqn:novelty_score}
    novelty = 
    \begin{cases}
    dist(x,0) & \text{if } len(d) = 0\\
    \min_{\forall s \in d} dist(x,s) ,  & \text{otherwise}
    \end{cases}
\end{equation*}

This method differs from the Diversity heuristic  by explicitly trying to find new individual prints that do not cover the same users as previous prints. Therefore, if the same users that are already covered in the dictionary are found in a generated print, the print's novelty score will decrease. Individually, these prints could have a lower coverage percentage than an individual Diversity MasterPrints but compensate in performance by covering a completely different group of users in the user space. 

\noindent{\bf{Distinguishing Aspects:}} While both heuristics use diversity searches, the methods can have different purposes and use cases. The diversity heuristic could be used when the size of the user space is known in the attacking scope or when overlaps between users in a group of fingerprints would be less punishing. On the contrary, the novelty heuristic would be better at avoiding overlaps in users between prints and is more focused on finding as many new users as possible. This heuristic could be used in a situation where the user space is unbounded or unknown, or where user samples are more individually distinguishable. 

Both methods have the most impact when allowed to search for multiple prints. If the verification system only allowed one chance for a fingerprint match, then the DeepMasterPrints method could be the best to use. However, the DeepMasterPrints were allowed unlimited iterations to find as many users as possible from the evolution starting point. If there was a set time limit, the Diversity or Novelty Heuristic generated print with the most users matched from the set would most likely outperform the DeepMasterPrint in matchability. Regardless of the methodology of search used, both heuristics can encapsulate a match rate with a wider set of users than a singular DeepMasterPrint could - as shown in the results section of this paper. 

\section{Experiment Setup}
Like Bontrager's experiments, we use the same dataset with its associated preprocessing steps for training and testing our generative models: the capacitive fingerprints from the FingerPass DB7 dataset \cite{jia2012cross}. Each fingerprint was reduced to a 128x128 grayscale image by removing whitespace from the already partial print for the capacitive dataset. 
We use the commercial fingerprint SDK, VeriFinger 12.0 as the target for our experiments. Version mismatch unfortunately makes a direct comparison to Bontrager's results impossible.

To keep as consistent as possible with Bontrager's work, we attempted to create a new Wasserstein GAN (WGAN) to generate fingerprints from the evolved latent vectors. Initial experiments with this WGAN were unsatisfactory so instead, we used a variational autoencoder as our generator for these experiments as these produced better results overall. The variational autoencoder (VAE) was trained on reconstruction of a real fingerprint image rather than adversarially as the WGAN would. The fingerprint image was passed to the encoder to be decoded to a latent vector, then passed to the decoder to recreate the fingerprint. By using the decoder network as the generator, we were able to input an evolved latent vector and receive an artificial fingerprint like the generator of the WGAN. 

The dataset was split in half for training and testing in the same manner as in Bontrager's work. None of the users in the training set are in the testing set and the sets are thus exclusive. 
10 trials were used for each experiment, and a maximum dictionary size of 10 was allowed for the Diversity and Novelty MasterPrints experiments, resulting 100 prints. For fairness, the novel approaches used 1000 generations for each print and the DeepMasterPrint approach used 10000 generations.

\section{Results}

The tables show the averaged trial results' coverages for the DeepMasterPrints (D), Diversity MasterPrints (I), Novelty MasterPrints (N), and randomly generated prints as a baseline (R) trained on the Capacitive fingerprint dataset using the VeriFinger SDK (VSDK).

\begin{table}[]
\small
\centering
\begin{tabular}{|ll|l|l|l|l|}
\hline
\multicolumn{2}{|l|}{\textit{VeriFinger Classifer}}     & \multicolumn{1}{c|}{R}       & \multicolumn{1}{c|}{D} & \multicolumn{1}{c|}{I} & \multicolumn{1}{c|}{N} \\ \hline
\multicolumn{1}{|l|}{\multirow{2}{*}{FMR 1.0}}  & Train & \multicolumn{1}{r|}{52.66} & 78.83                & \textbf{96.75}       & 95.97                \\ \cline{2-6} 
\multicolumn{1}{|l|}{}                          & Test  & \multicolumn{1}{r|}{53.77} & 72.72                & 93.66                & \textbf{96.69}       \\ \hline
\multicolumn{1}{|l|}{\multirow{2}{*}{FMR 0.1}}  & Train & \multicolumn{1}{r|}{7.63}  & 25.93                & 47.30                & \textbf{48.25}       \\ \cline{2-6} 
\multicolumn{1}{|l|}{}                          & Test  & 6.75                       & 17.92                & 33.11                & \textbf{40.86}       \\ \hline
\multicolumn{1}{|l|}{\multirow{2}{*}{FMR 0.01}} & Train & 0.58                       & 3.63                 & \textbf{10.39}       & 10.19                \\ \cline{2-6} 
\multicolumn{1}{|l|}{}                          & Test  & 0.14                       & 0.95                 & \textbf{2.52}        & 2.38                 \\ \hline
\end{tabular}
\caption{Average FMR match percentages for each experiment type: [R]andom, [D]eep Master Prints, D[I]versity, [N]ovelty on the VSDK}
\label{label:VeriFinger_res}
\end{table}

The baseline experiments produced 10 random latent vectors (to match the 10 prints in the dictionary) and generated fingerprints from the VAE. These "random" prints - even when unevolved - still managed to look like realistic fingerprints and match with a few users from the VeriFinger classifier. This outperforms Bontrager's Wasserstein GAN which reported to have no matches from the VeriFinger classifier when unevolved randomized latent vectors were passed to the generator \cite{bontrager2018deepmasterprints}. Thus, the VAE is capable of producing higher quality fingerprints that are able to match with users in the database regardless of the input latent vector. The dictionary of 10 random prints were able to match more than half of the users from the training set using an FMR rate of 1\%, and slightly more from the test set. Evaluating with FMR rates of 0.1\% and 0.01\% produced 7\% and 0.1\% user matches respectively. 

\begin{figure}[hbt]
    \centering
    \includegraphics[width=0.9\linewidth]{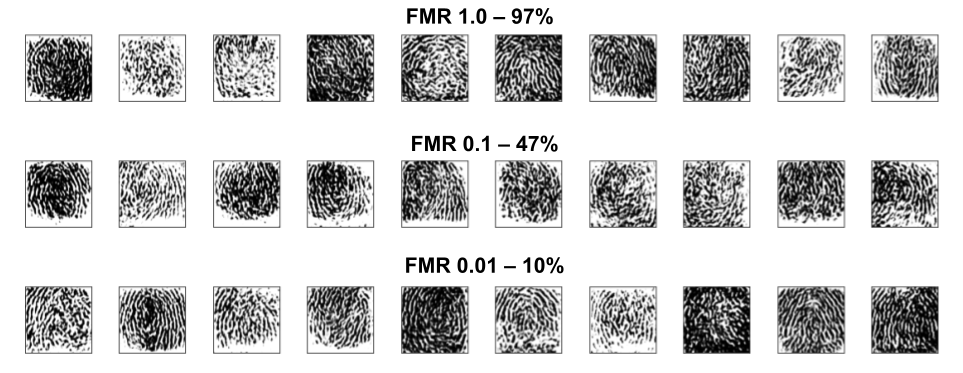}
    \caption{Dictionary of Diversity MasterPrints with the most coverage per FMR rate (Verifinger)}
    \label{fig:v_exp2_prints}
\end{figure}

The match results from the DeepMasterPrints experiments using the VAE were on-par with Bontrager's original results using the Wasserstein GAN. His work allowed the fingerprints to evolve until the change in fitness value plateaued - which led to an unspecified bound for the number of evolution iterations. Our DeepMasterPrints experiments had a maximum of 10k iterations, which could explain the slight decrease in matches. However, the DeepMasterPrints had an average match rate of around 78\%, 25\%, and 3\% for the FMR rates of 1\%, 0.1\%, and 0.01\% respectively on the training dataset. These prints did not generalize as well as the randomized prints and had a lower match rate on the test set. The DeepMasterPrints with the most coverage from the trials can be seen in figure \ref{fig:v_exp2_prints}.

\begin{figure}[hbt]
    \centering
    \includegraphics[width=0.9\linewidth]{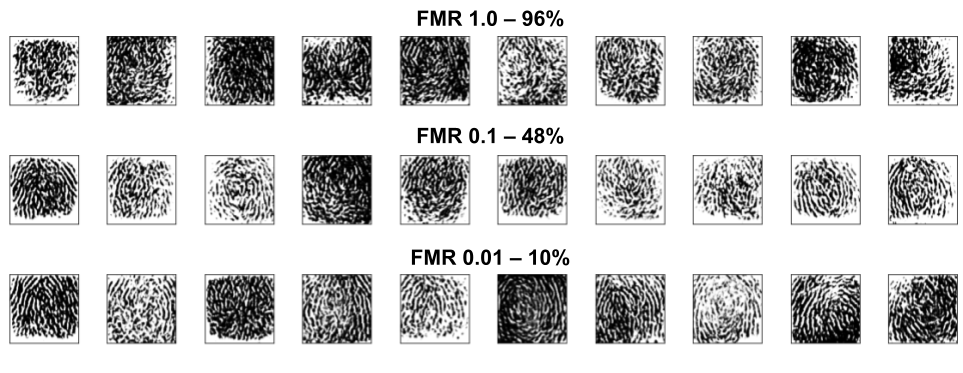}
    \caption{Dictionary of Novelty MasterPrints with the most coverage per FMR rate (Verifinger)}
    \label{fig:v_exp3_prints}
\end{figure}

The Diversity and Novelty MasterPrints both greatly outperformed the random prints and the DeepMasterPrints - even while the DeepMasterPrints were allowed to evolve for 10 times as many iterations. At an FMR rate of 1.0\% both experiments were able to match almost over 95\% of training dataset and generalize to the testing dataset to match over 90\%. In the case of the Novelty MasterPrints, more average matches were found in the testing set than the training set. Even at lower FMR rates, both prints managed to get almost double and triple the matches in the training set compared to the DeepMasterPrints and still maintain generalizability to the testing set. 
The dictionaries with the most coverage Diversity and Novelty MasterPrints can be seen in figures \ref{fig:v_exp2_prints} and \ref{fig:v_exp3_prints} respectively.

\section{Conclusion}
This work demonstrates the efficacy of optimizing for matching multiple users in a biometric system via a evolutionary diversity search. By creating a set of generated fingerprints with each new print attempting to cover the remaining subset of users, the match rate drastically increases over using a singular generated print that is evolved for the same amount of time, and also over multiple prints that are evolved independently. While both of the novel methods perform well, novelty prints have the overall best performance. 
While this system has only been applied to fingerprints, this pipeline could also be applied to other biometric verification systems such as faces or voices.
We also want to explore quality diversity algorithms \cite{pugh2016quality} to further expand the possibility space for biometric attacks. 

\bibliographystyle{plain}
\bibliography{references}

\end{document}